%% file: main.tex
\documentclass[10pt]{article} 
\usepackage[preprint]{tmlr}

\input{math_commands.tex}

\usepackage{hyperref}
\usepackage{url}
\usepackage[utf8]{inputenc} 
\usepackage[T1]{fontenc}    
\usepackage{hyperref}       
\usepackage{url}            
\usepackage{booktabs}       
\usepackage{amsfonts}       
\usepackage{nicefrac}       
\usepackage{microtype}      
\usepackage{xcolor}         
\usepackage{amsmath} 
\usepackage{xcolor}
\usepackage{soul}
\usepackage{multirow}
\usepackage{amssymb}

\usepackage{times}
\usepackage{latexsym}
\usepackage{algorithm, algorithmic}
\usepackage{graphicx}
\usepackage{subcaption}
\usepackage{tabularx}
\usepackage{enumitem}
\usepackage{svg}
\usepackage{tabularray}
\usepackage{longtable}
\usepackage{amsthm}
\usepackage{bm}
\usepackage{stackengine}

\usepackage{cleveref}

\definecolor{mycyan}{RGB}{150, 230, 255} 

\setitemize{labelindent=2em,labelsep=0.2cm,leftmargin=*}
\title{What Am I Missing? Question-Answering as Hidden State Probing}

\author{
\name Chu Fei Luo \email chufei.luo@queensu.ca \\
\addr Department of Electrical and Computer Engineering \& Ingenuity Labs, Queen's University \\
Conflict Analytics Lab, Queen's University \\
Vector Institute for AI
\AND
\name Samuel Dahan \email samuel.dahan@queensu.ca \\
\addr Conflict Analytics Lab, Queen's University \\
Cornell Law School
\AND
\name Xiaodan Zhu \email xiaodan.zhu@queensu.ca \\
\addr Department of Electrical and Computer Engineering \& Ingenuity Labs, Queen's University \\
Conflict Analytics Lab, Queen's University \\
Vector Institute for AI
}

\def\month{MM}  
\def\year{YYYY} 
\def\openreview{\url{https://openreview.net/forum?id=XXXX}} 

\begin{document}

\maketitle
\begin{abstract}




The study of test-time reasoning in large language models (LLMs) has attracted increasing interests due to its profound impact on many tasks, with early work tracing back to approaches such as chain-of-thought reasoning. The basic mechanism underlying the test-time reasoning process, however, is still under-studied: from the same input prompt, and even the same partial solution, LLMs can produce varied answers if sampled multiple times. In this paper,
we leverage question-asking as an inference-time intervention that articulates information about the model's hidden state. Specifically, we present a student-teacher setting where a student asks questions to a teacher. We train a probe on the student's hidden state before and after asking a question and find it is predictive of the trajectory's final correctness, even before generating the teacher's answer. This suggests there is a meaningful signal from the \emph{self-diagnosis} that occurs during question generation rather than information transfer from the teacher. We then frame question-asking as a sequential decision problem, using this probe as a quality score, and define a gating policy to ask questions that maximize likelihood of correctness. We find that the success of question-asking as an intervention is largely dependent on the model's self-consistency. Our empirical results show a gap between detection and recovery; while our gating policy captures model correctness and uncertainty, interventions are equally likely to harm correct trajectories as they are to recover incorrect ones. This gap between diagnosis and correction has broader implications on language models' capacity for self-refinement under uncertainty.

\footnote{Our code will be made available at [placeholder] upon acceptance.}

\end{abstract}
\section{Introduction}
Since the introduction of chain-of-thought reasoning, test-time technologies have become a significant field of study for large language models (LLMs). This has also led to the introduction of Large Reasoning Models (LRMs) that undergo post-training to encourage the model to produce intermediate tokens before formulating the final answer \cite{kambhampati2025stop}. There have been various interventions to improve LLMs' performance at inference time by manipulating intermediate reasoning tokens \cite{ wei2022chain, snell2024scaling, zhang2025survey}. 
For example, \citet{muennighoff-etal-2025-s1} demonstrate that simply replacing the stop token with a continuation such as ``wait,'' often leads to significant improvements in accuracy. In addition, other works treat language model generation as an exploration in the generation space and employ branching to search for optimal reasoning trajectories \cite{li-etal-2026-entropy, huang2025best, chang2025step}. The third class of intervention is through question-asking (QA) --- questions are an effective mechanism for clarification \cite{aliannejadi2019asking, rao-daume-iii-2018-learning}, gathering new information \cite{hu2024uncertainty}, or encouraging the decomposition of complex tasks \cite{press2023measuring}. These interventions aimed to compensate for the same fundamental issue: LLMs struggle to consistently self-correct reasoning without external feedback \cite{huang2023large, tyen2024llms}.


 In this paper, we investigate question-asking as a structured inference-time intervention for both diagnosing and correcting test-time reasoning. A central challenge in self-correction is that LLMs often fail to identify when their intermediate reasoning has gone wrong without external feedback. Question generation provides a natural mechanism for addressing this challenge: to formulate a useful question, the model must articulate what is missing, uncertain, or unresolved in its current reasoning state. Because such questions can encode semantically targeted information~\cite{ambati2025socratic}, we hypothesize that the act of asking a question can both surface latent signals about trajectory quality and induce a meaningful change in the subsequent reasoning trajectory, potentially steering the model toward the correct solution.
We propose two specific hypotheses: First, if the reasoning tokens are meaningful, then introducing new semantic information would help steer the trajectory. Second, if a model is truly ``proficient'' at reasoning, it would be able to ask questions that can correct its own logic.
The objective of this study is to leverage question-asking to predict the correctness of the reasoning trajectory, and to introduce interventions to question and recover from uncertain steps. 
We frame our study as a decision-making problem in a student-teacher setup, where a student is asking questions to a teacher and re-incorporating the teacher's answer into its own reasoning trajectory. We introduce a gating policy that uses hidden-state probing to assess two axes of good questions: (i) \textit{which to ask}: what student questions would lead to the most fruitful outcomes? and (ii) \textit{when to ask}: when is an intervention likely beneficial to the student?

Our contributions are summarized as follows:
\begin{itemize}
    \item We demonstrate that prompting a reasoning model to generate questions at test-time produces a measurable shift in the hidden state \textit{before} incorporating the answer. We train a hidden state probe and empirically demonstrate the probe's scoring is reasonably well-calibrated to the final correctness rate. Additionally, we observe the \textit{range} of predicted correctness is well-correlated to correctness rate, more than per-set semantic diversity.
    \item We frame question-asking as a sequential decision problem in a student-teacher setting and propose a pipeline with a gating mechanism based on our probe-estimated correctness, to intervene on a reasoning trajectory when it is likely to be wrong. This pipeline frames question-asking as trajectory recovery, where we need to shift incorrect answer distributions towards a desired direction.
    \item We present thorough ablations to analyze the mechanisms of question-asking and find, despite identifying incorrect trajectories, there exists a gap between detection and recovery. We show that the success of an intervention is strongly correlated to baseline self-consistency. Even though we can predict correctness from the hidden state, the student does not necessarily utilize this signal to assess its own trajectory.
\end{itemize}

\section{Related work}
\begin{figure}
    \centering
    \includegraphics[width=0.9\linewidth, trim=0.5cm 0.5cm 0.5cm 0.25cm, clip]{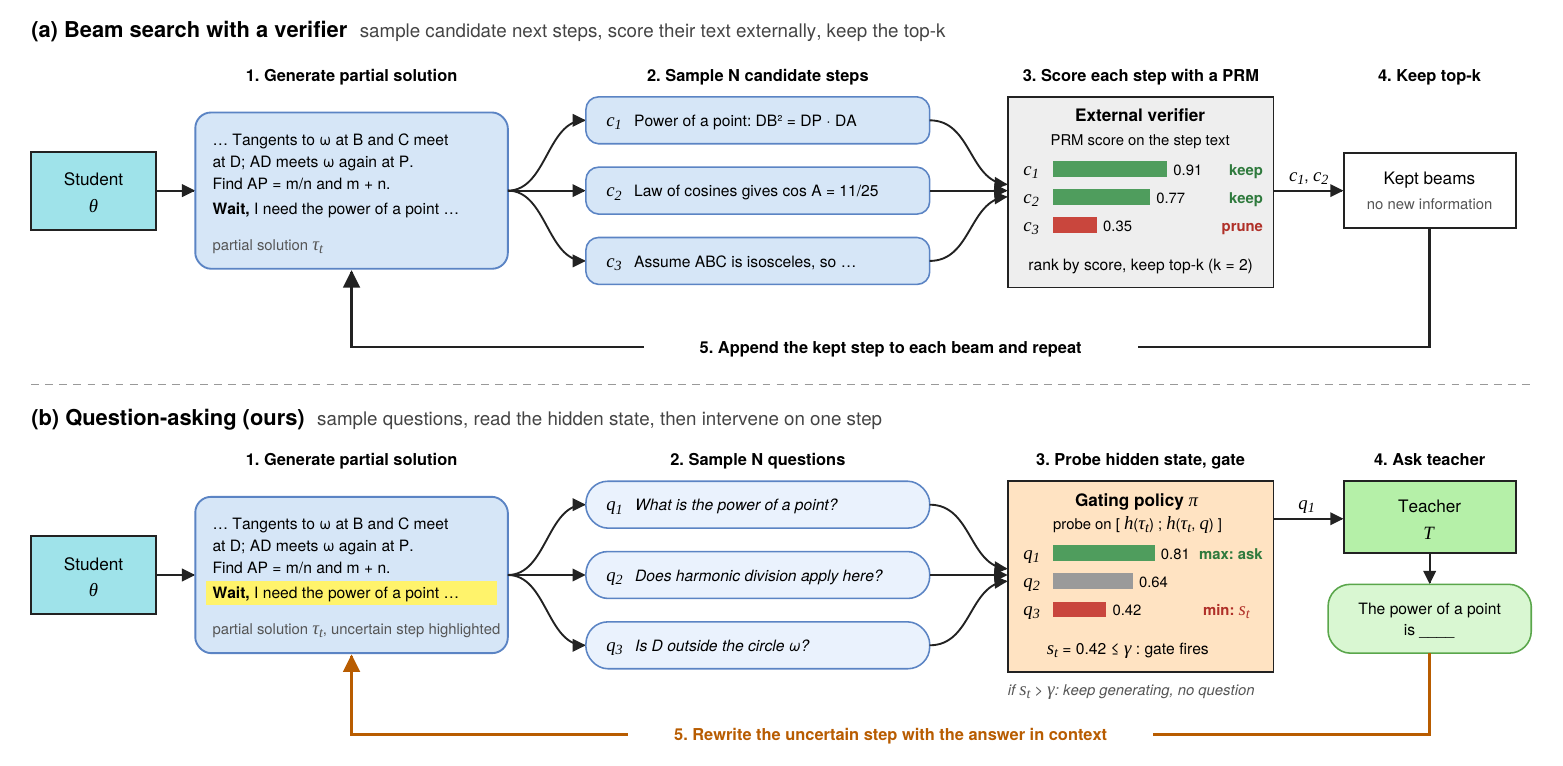}
    \caption{An illustration of our experimental setting. A student $\theta$ generates a partial solution $\tau$ for a target task. First, it chooses a step in the reasoning trajectory to pause the generation. At scheduled intervals determined by rule-based or external signal, we check a gating policy $\pi$. We sample $n$ questions from the student and choose the best to be sent to the teacher $T$. After answering, the student incorporates the answer into a rewrite of the uncertain step.}
    \label{fig:pipeline}
\end{figure}
\input{sections/related_work}

\section{Methodology}
\label{sec:method}

\input{sections/methodology-v2}

\vspace{-0.5em}
\section{Experimental Settings}
\label{sec:exp}

\vspace{-0.5em}
\paragraph{Models}

We use Qwen3 \cite{yang2025qwen3} at 1.7B, 4B, and 14B parameters. For the PRM, we use Qwen2.5-Math-PRM-7B \cite{zhang2025lessons}. We theorize Qwen3 models are well-calibrated to math correctness, and have a stronger signal to be surfaced from their hidden state. To analyze the transferability to another model family, we also include results on Olmo-3-7B-Think \cite{olmo2025olmo}.

We mostly test math reasoning to study the correlation between proficiency and recovery --- Qwen3 models already achieve strong baseline performance on these datasets, which allows us to test self-correction on known capability. Also, process reward models are more calibrated on verifiable reasoning, so they are an effective external signal. To focus on the model's capacity for self-correction, most of our study is self-play --- i.e. the same model plays the role of both student and teacher with the same available information.

\vspace{-0.5em}
\paragraph{Datasets}
We evaluate on three math datasets:
\begin{itemize}
    \item GSM8k~\cite{cobbe2021gsm8k} --- 1{,}319 samples of grade school level word problems in math. We take 100 samples of the train split for training our hidden state probe. oh
    \item Math500~\cite{hendrycksmath2021} --- 500 samples of competition-level math. We take the first 15 samples as a training set for the hidden state probe, and report results on the 485 remaining samples.
    \item AIME24~\cite{aime24} --- A held-out set of 30 Math Olympiad questions from 2024.
\end{itemize}

\vspace{-0.5em}
\paragraph{Hyperparameters}
For our gating threshold, we choose $\gamma$ to be 0.6 after a hyperparameter sweep. We set $n$ to be 5 reasoning steps (split by newlines, not individual sentences), and $m$ to be 5, as we find that there are minimal gains beyond 5 interventions.
Unless otherwise specified, the default metric is Majority Vote accuracy over $K=16$ rollouts. Since we report majority vote, we use bootstrapping to approximate confidence intervals in our results. Please see Appendix \ref{app:exp} for more details.

\vspace{-0.5em}
\section{Results}
\label{sec:results}
\vspace{-0.5em}
\input{sections/results}

\input{sections/rebuttal}
\vspace{-0.5em}
\section{Conclusion}

\vspace{-0.75em}
This work investigates the dynamics of question-asking as a test-time intervention. 
Our study shows that 
(1) models partially encode their own correctness in the hidden state shift induced by asking a question; 
(2) this signal is weakly correlated to difficulty in a way that external PRM signals are not; and
(3) there is a gap between detecting this signal and recovering the trajectory.
While interrupting a reasoning model with a question prompt can change the overall trajectory, there is no consistent shift towards correctness. Instead, we discover that the rate of distribution shift is more correlated with the model's confidence, where targeting uncertainty is also likely to cause the model to second-guess correct trajectories.
We demonstrate that comparing logprob-based selection to our probe-based scoring, the best question is often not the most probable one. These results demonstrate a gap between hidden state detection and recovery in the reasoning trace. LLMs still struggle to identify flaws in their own logic, even though the signal exists in the hidden state. 
This gap has broader implications on self-refinement, especially when the model is uncertain, and we encourage further research to improve the model's inherent ability to self-diagnose or understand its own actions.

\newpage



\bibliographystyle{abbrvnat}
\bibliography{custom}

\newpage
\include{sections/appendix}

\end{document}

%% file: math_commands.tex
\usepackage{amsmath,amsfonts,bm}

\def\eqref#1{equation~\ref{#1}}

\def\1{\bm{1}}

\DeclareMathAlphabet{\mathsfit}{\encodingdefault}{\sfdefault}{m}{sl}
\SetMathAlphabet{\mathsfit}{bold}{\encodingdefault}{\sfdefault}{bx}{n}



%% file: sections/related_work.tex
\paragraph{Self-correction and self-refinement.}
There is a large body of work that studies whether language models can improve their own reasoning through iterative refinement \cite{madaan2023self} or, more broadly, self-evolution \cite{tao2024survey}. \citet{wang2022self} propose \emph{self-consistency}, which samples multiple chains-of-thought (CoTs) and aggregates via majority vote. 
This provides a critical baseline for test-time compute: if repeated sampling achieves the same gains as interaction, external feedback may be unnecessary. 
\citet{zelikman2022star} introduce STaR, where models iteratively refine reasoning using hints derived from correct answers. Similarly, \citet{shinn2023reflexion} propose Reflexion, where models critique failed attempts and retry using feedback signals. 
These approaches demonstrate that retrying with additional information can improve performance. 
However, they rely on oracle signals (e.g., correctness labels or ground-truth answers) to guide refinement. \citet{huang2023large} and \citet{tyen2024llms} demonstrate that LLMs struggle with self-correction in the absence of external feedback. We investigate whether there are internal signals of correctness, and whether or not they translate into action.


\paragraph{Asking clarifying questions.}
Several works study whether models can improve reasoning by asking follow-up questions. Self-Ask \citep{press2023measuring} prompts a model to decompose problems into sub-questions that it asks itself, or feeds into an external search engine. Prior work in dialogue and information retrieval \citep{rao-daume-iii-2018-learning, aliannejadi2019asking} studies when clarification questions are worth asking, framing the problem as a trade-off between cost and expected utility.
Our work differs in two key ways. First, we explicitly separate the \emph{student} (reasoner) and \emph{teacher} (answering model), allowing us to analyze the effect of interaction. 
Second, we evaluate whether asking questions provides value beyond simply triggering regeneration. 

\paragraph{Internal representations and probing.}
There are many works that study what information is encoded in neural representations via probing classifiers \citep{belinkov-2022-probing, wang-xu-2025-thoughtprobe, press2023measuring}. \citet{zhu2025llm} finds that it is possible to predict the final correctness from the input question with a linear probe, which they use to inform their budget for test-time scaling. In contrast, we use the hidden state shift over a question and provide detailed analysis of the actual utility of our probe at inference time.


%% file: sections/methodology-v2.tex

\subsection{Setup}

Our setup is illustrated in \Cref{fig:pipeline}. We frame our study as a sequential decision problem, where a student is gated by an external policy to ask questions to a teacher. Let $x$ denote a problem instance and $y \in \mathcal{Y}$ a discrete answer. A student model $\theta$ induces a distribution $p_\theta(y \mid x)$ over answers by sampling a full rollout. A question $q$ posed to a teacher $p_T$ produces a response $a$, after which the student updates its answer distribution to $p_\theta(y \mid x, q, a)$.

The intuitive idea is simple; the student only asks questions when the reasoning trajectory it has generated is . We wish to find a score $s_t$ that measures the expected quality of the current trajectory. At a time step $t$, we decide whether to trigger the intervention. Let $\tau_t = (x, z_{1:t})$ denote the partial reasoning trajectory. We intervene on the partial reasoning trajectory with a gating policy $\pi(\tau_t)$, formally defined as the following threshold decision at step $t$:

\begin{equation}
\label{eq:gate}
\pi(\tau_t) =
\begin{cases}
\pi(\cdot \mid \tau_t), & s_t \le \gamma \\
\varnothing, & \text{otherwise}.
\end{cases}
\end{equation}





\subsection{Probing Test-time Question-Asking}

Using this definition of question-asking as a decision made over a partial trajectory $\tau_t$, we seek to predict the effectiveness of question-asking as a test-time intervention strategy.
We study the value of the question itself, and also the timing of the question --- that is, does there exist a critical point where a trajectory can be rescued and guided towards correctness. Empirically, we find that we can predict $p_\theta(y \mid x, q, a)$ using a probe trained on the hidden state shift induced by the question $q$. 
Using this probe, we can approximate the correctness of the current reasoning trajectory. We utilize this correctness score to decide whether or not to ask a question, formulated as a sequential decision problem. 

\paragraph{Belief State Probe.} We measure the belief state as the answer distribution obtained by sampling full rollouts from a reasoning trajectory. For a partial reasoning trajectory $\tau_t$, the student maintains a distribution $Y_t$ over the answer space obtained by sampling continuations, i.e.
$
Y_t := p_\theta(y \mid \tau_t).
$
At time $t$, the student samples a question $q_t$ from a feasible set $\mathcal{Q}_\theta(\tau_t)$, and updates the context to the triplet $(\tau_t, q_t, a_t)$. 
The answer distribution after a QA intervention $Y_q$ is defined as:
\[
Y_q := p_\theta(y \mid \tau_t, q_t, a_t).
\]

We train a hidden state probe to predict the correctness rate at time step $t$ \emph{before generating the teacher answer}. 
That is, for a hidden state $h_\theta$, we train a probe to predict the correctness rate $P(y_q = y\star)$ where $y_q \sim Y_q$. We take the hidden state as the embedding of the last token at the final layer of the model, i.e. the same embedding used for token prediction. We concatenate the hidden states $h_\theta(\tau_t)$ and $ h_\theta(\tau_t,q_t)$, before and after asking the question.


\paragraph{Question-Asking as a Test-time Intervention} 
We formulate what to ask as the question that maximizes the expected intervention outcome, or maximally increases $\Delta_{acc}$. Next, we discuss the timing --- when to ask --- as a decision made over the belief state $\tau$. Unlike \Cref{eq:acc}, we assume no access to the baseline answer distribution $\mathbb{P}_{y \sim p_\theta(\cdot \mid x)}(y = y^\star).$ We sample $N$ discrete questions $Q(\tau_t)$, and take the score that triggers the gating policy $\pi(\tau_t)$ as the following:
\begin{equation}
s_t = \min_{q \in Q(\tau_t)} \mathbb{E}{y \sim p_\theta(\cdot \mid \tau_t, q)}[\mathbf{1}[y = y^\star]]
\end{equation}


Note the minimum term --- this again derives from the idea of treating a hidden state as a belief state, and asking questions is an articulation of possible trajectories that could occur from a specific reasoning trajectory. If there is at least one question sample that is predicted to lead to a weak belief state, then the gate triggers. This is intuitively equating model uncertainty to trajectory variance; that is, if a bad final state is reachable from the current reasoning step, then that can be treated as information on the trajectory as a whole. We then choose the \textbf{maximum}-scoring question from the sampled set to ask the teacher, the teacher answers, and we \emph{rewrite} the step triggered with the teacher's answer in context.


\paragraph{Gate Scheduling} Another benefit of studying math problems is that there are many well-trained process reward models (PRMs) that provide an external source of correctness. 
We test several timing strategies to decide \textit{when} to check the gating policy:
\begin{itemize}
    \item \textbf{Fixed interval:} Always trigger the gate check every $n$ reasoning steps. Additionally, we run $m$ iterations of the QA loop and keep the final answer at iteration $m$.
    \item \textbf{Adaptive} (PRM-weak): Only trigger the gating decision on steps where the PRM score falls below a threshold $\alpha$. We additionally continue to employ the idea of answer mass as an early stopping criteria --- that is, if all sampled trajectories exceed the threshold and there is low variance between the samples, we assume sufficiency and stop.
\end{itemize}

\subsection{Measuring Intervention Quality}
\label{sec:d_acc}

We study the effect of test-time interaction through the change in the student's answer distribution.
This can be conceptualized as a Jensen-Shannon divergence between the student's distribution in the answer space pre- and post- question, i.e. 
$\Delta_q(x) \;=\; D_{\mathrm{JS}}\big(p_\theta(y \mid x) \,\|\, p_\theta(y \mid x, q, a)\big)$. We refer to $\Delta_q(x)$ as the \emph{distributional shift} induced by question $q$.

The quantity $\Delta_q(x)$ captures the magnitude of change, but not whether the change improves correctness. One primary goal of this study is to assess the direction of the change in trajectory induced by a question $q$, such that we can distinguish and characterize helpful from harmful questions. Let $y^\star$ denote the correct answer.

We define the expected accuracy change induced by a question and a sampled answer as:
\begin{equation}
\Delta_{\mathrm{acc}}(q;x)
=
\mathbb{P}_{y \sim p_\theta(\cdot \mid x,q,a)}(y = y^\star)
-
\mathbb{P}_{y \sim p_\theta(\cdot \mid x)}(y = y^\star)
\end{equation}

In practice, we estimate this quantity via $K$ discrete rollouts sampled over the reasoning trajectory with and without the question:
\begin{equation}
\hat{\Delta}_{\mathrm{acc}}(q;x)
=
\frac{1}{K} \sum_{i=1}^{K} \mathbf{1}[y_{i,q} = y^\star]
-
\frac{1}{K} \sum_{i=1}^{K} \mathbf{1}[y_i = y^\star]
\label{eq:acc}
\end{equation}
where $y_{i,q} \sim p_\theta(\cdot \mid x,q,a)$ and $y_i \sim p_\theta(\cdot \mid x)$.

An intervention by our pipeline is \textit{helpful} if $\Delta_{\mathrm{acc}}(q;x) > 0$ and \textit{harmful} if $\Delta_{\mathrm{acc}} < 0$. The expected accuracy gain $\Delta_{\mathrm{acc}}(q;x)$ involves two sources of stochasticity: (i) the teacher response $a \sim  p_T(\cdot \mid x,q)$ and (ii) the student’s decoding $y \sim p_\theta(\cdot \mid x,q,a)$. 
This $\Delta_{acc}$ formulation is meant to capture the benefit of question-asking compared to the counterfactual (no intervention), but it is measured post-hoc. We do not assume access to the baseline answer at inference time. In other words, an optimal intervention is one that maximizes this shift in the answer distribution, not necessarily one where the right answer is reachable but rarely sampled. We measure this by reporting majority vote accuracy.

%% file: sections/results.tex

\begin{table}[t]
\centering
\caption{Our main results as compared to a zero-shot baseline. (nt) indicates Non-thinking and (t) indicates Thinking.
We report Mean Questions as questions asked to the teacher, accuracy under majority vote over 5 samples (maj@5), and mean tokens per generation.
\textit{Adaptive}: PRM-weak steps and early stopping; \textit{Fixed}: fixed interval. Subscripts report the 95\% confidence interval.}
\label{tab:belief-cost-per-model}
\small
\setlength{\tabcolsep}{3pt}
\resizebox{\linewidth}{!}{%
\begin{tabular}{@{}|c|lc |rrr | rrr |rrr@{}}
\toprule
& & &\multicolumn{3}{c}{\textbf{Accuracy, maj@5 (\%)}}  & \multicolumn{3}{c}{\textbf{Mean Questions}} & \multicolumn{3}{c}{\textbf{Mean Tokens}} \\
\cmidrule(lr){4-6} \cmidrule(lr){7-9} \cmidrule(lr){10-12}
\textbf{Model} & \textbf{Setting} & \textbf{Mode} & \textbf{GSM} & \textbf{MATH} & \textbf{AIME} & \textbf{GSM} & \textbf{MATH} & \textbf{AIME} & \textbf{GSM} & \textbf{MATH} & \textbf{AIME} \\
\midrule
  \multirow{4}{*}{\rotatebox{90}{Qwen3-1.7B}}
  & Baseline &(nt) & 84.9$_{[-2.0,\,+2.0]}$ & 76.5$_{[-3.9,\,+3.9]}$ & 16.7$_{[-13.4,\,+13.3]}$ & --- & --- & --- & 265 & 871 & 2{,}933 \\
     & Fixed& (nt) & 84.3$_{[-1.9,\,+2.0]}$ & 75.9$_{[-3.7,\,+3.7]}$ & 13.3$_{[-10.0,\,+13.4]}$ & 0.3 & 2.1 & 3.8 & 1{,}356 & 5{,}391 & 16{,}772 \\ 

     & Adaptive& (nt) & 86.1$_{[-1.9,\,+1.9]}$ & 76.9$_{[-3.7,\,+3.9]}$ & 13.3$_{[-10.0,\,+13.4]}$ & 0.1 & 0.5 & 1.2 & 318 & 1{,}260 & 4{,}584 \\

     & Baseline &(t) & 91.4$_{[-1.4,\,+1.5]}$ & 92.8$_{[-2.5,\,+2.3]}$ & 55.6$_{[-14.1,\,+14.2]}$ & --- & --- & --- & 1{,}380 & 5{,}074 & 13{,}570 \\
     & Adaptive& (t) & 91.6$_{[-3.1,\,+2.8]}$ & 92.0$_{[-2.3,\,+2.2]}$ & 56.8$_{[-15.3,\,+14.8]}$ & 0.1 & 1.0 & 2.0 & 840 & 4{,}932 & 14{,}388 \\


\midrule
  
  \multirow{4}{*}{\rotatebox{90}{Qwen3-4B}} & Baseline &(nt) & 90.6$_{[-2.1,\,+1.9]}$ & 86.6$_{[-3.1,\,+2.9]}$ & 36.7$_{[-16.7,\,+16.6]}$ & --- & --- & --- & 270 & 698 &  1{,}271 \\
     & Fixed& (nt) & 92.6$_{[-1.5,\,+1.3]}$ & 86.0$_{[-2.3,\,+3.7]}$ & 28.6$_{[-14.3,\,+17.8]}$  & 0.0 & 1.0 & 2.3 & 1{,}374 & 4{,}885 & 20{,}523 \\
     & Adaptive& (nt) & 93.1$_{[-1.4,\,+1.4]}$ & 85.6$_{[-2.6,\,+3.2]}$ & 30.0$_{[-16.7,\,+16.7]}$  & 0.0 & 0.2 & 1.0 & 296 & 1{,}104 & 5{,}317 \\
     & Baseline &(t) & 95.2$_{[-1.2,\,+1.2]}$ & 96.7$_{[-1.8,\,+1.8]}$ & 76.7$_{[-16.7,\,+13.3]}$ & --- & --- & --- & 2{,}327 & 6{,}897 & 12{,}259 \\
     & Adaptive& (t) & 96.1$_{[-2.1,\,+0.3]}$ & 95.8$_{[-1.8,\,+1.6]}$ & 80.8$_{[-14.4,\,+12.2]}$ &  0.0 & 0.6 & 0.7 & 568 & 3{,}207 & 5{,}863 \\

\midrule
  \multirow{4}{*}{\rotatebox{90}{Qwen3-14B}}
    & Baseline& (nt) & 96.0$_{[-2.2,\,+2.0]}$ & 88.5$_{[-3.1,\,+3.1]}$ & 43.3$_{[-16.6,\,+16.7]}$ & --- & --- & --- & 192 & 840 & 4{,}324 \\
     & Fixed& (nt)& 95.2$_{[-1.6,\,+1.3]}$ & 87.8$_{[-3.1,\,+2.7]}$ & 40.0$_{[-16.7,\,+20.0]}$ & 0.1 & 3.9 & 3.1 & 1{,}115 & 4{,}058 & 17{,}730 \\
     & Adaptive& (nt) & 95.4$_{[-1.3,\,+1.1]}$ & 89.1$_{[-2.9,\,+2.7]}$ & 36.7$_{[-16.7,\,+16.6]}$ & 0.0 & 0.2 & 0.9 & 200 & 979 & 3{,}915 \\
     & Baseline& (t) & 96.6$_{[-1.1,\,+1.0]}$ & 97.1$_{[-2.9,\,+2.3]}$ & 80.0$_{[-30.0,\,+20.0]}$ & --- & --- & --- & 2{,}699 & 4{,}611 & 11{,}871 \\
     & Adaptive &(t) & 96.6$_{[-1.9,\,+1.8]}$ & 96.0$_{[-1.7,\,+1.5]}$ & 81.8$_{[-14.0,\,+12.3]}$  & 0.1 & 0.8 & 1.1 & 533 & 3{,}595 & 10{,}370 \\

\midrule
  \multirow{3}{*}{\rotatebox{90}{Olmo-3}}
     & Baseline &(t) & 94.9$_{[-2.5,\,+2.3]}$ & 91.2$_{[-4.1,\,+3.6]}$ & 76.7$_{[-16.7,\,+13.3]}$ & --- & --- & --- & 2{,}066 & 5{,}026 & 15{,}429 \\
     & Adaptive& (t) & 95.5$_{[-2.3,\,+2.0]}$ & 97.0$_{[-1.6,\,+1.4]}$ & 76.7$_{[-16.7,\,+13.3]}$  & 0.0 & 0.3 & 1.0 & 2{,}087 & 7{,}668 & 33{,}243 \\
     & Fixed &(t) & 95.7$_{[-2.2,\,+2.1]}$ & 97.3$_{[-1.9,\,+1.7]}$ & 73.3$_{[-16.6,\,+13.4]}$  & 0.3 & 1.0 & 2.0 & 10{,}103 & 31{,}212 & 93{,}340\\
\bottomrule
\end{tabular}
}

\end{table}

\subsection{Hidden state probe accuracy}

\vspace{-0.5em}

\paragraph{Correlation to actual correctness.} We take the predicted correctness rate from our probe in each model, and measure it against the actual correctness rate of the final rollouts. Our probe is reasonably calibrated to actual correctness; we achieve Brier scores of 0.157 for Qwen3-1.7B to 0.121 for Qwen3-14B \emph{without the teacher answer}. The calibration improves linearly with model size; the full list of scores along with p-value significance can be found in Appendix \ref{app:extra-tabs}. We tested various formulations --- including just the pre-question embedding $h_\theta(\tau_t)$ or just the post-question embedding $h_\theta(\tau_t,q)$ --- but the combination of pre+post was the most predictive of final correctness rate. This exemplifies the concept of hidden state shift; the question itself induces a change in the hidden state representation that is indicative of final performance.

\begin{figure}
      \centering
      \includegraphics[width=0.75\linewidth, trim={0.5cm 0.5cm 0.5cm 0cm},clip]{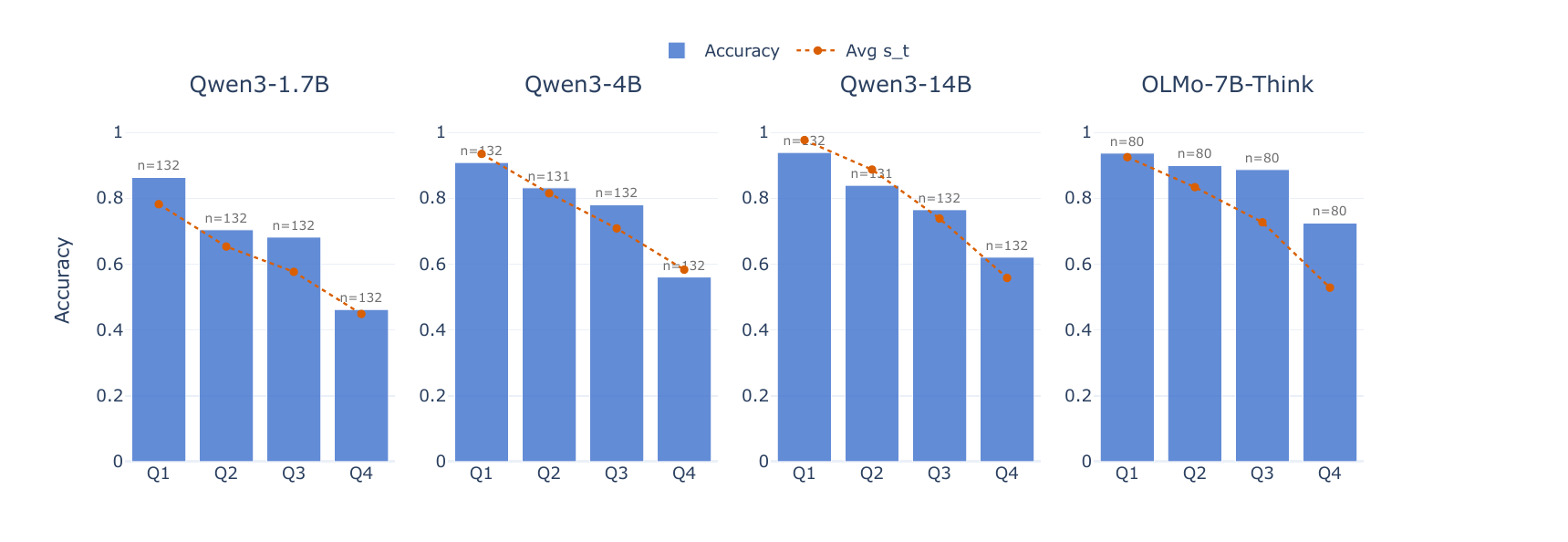}
      \caption{Majority vote accuracy vs. probe score range quartile across all belief-gated settings from \Cref{tab:belief-cost-per-model}, stratified by model. $s_t$ indicates the minimum predicted probe score, which we use to gate the QA intervention.}
      \label{fig:spread}
  \end{figure}

\vspace{-1em}
\paragraph{Hidden state probe as a predictor.}
For our fixed setting, we track the range of predicted probe scores as the diff between max. and min. scored correctness. The results are available in \Cref{fig:spread}. Binning the result into four quartiles, we observe that accuracy is negatively correlated with range; that is, the wider the range of sampled values, the lower the final model accuracy. This indicates that answer mass is detectable from the questions asked by the model, even before incorporating the teacher's answer, and this is strongly correlated with the student's final correctness.

\vspace{-0.5em}
\subsection{Effectiveness of Question-Asking Intervention}
\vspace{-0.5em}
Next, we use this hidden state probe score as a threshold value for our gating policy $\pi(\tau_t)$ in an LLM's reasoning trajectory. We seek to understand if this signal can be translated into a test-time intervention that can consistently detect and correct weaker trajectories. Our main findings are summarized in \Cref{tab:belief-cost-per-model}. maj@16 is reported for the baselines and Adaptive (t) on MATH and AIME; all other cells only report maj@5 due to computational constraints. 
We hypothesize that GSM8k's problem set is more suited to question-asking; there is, for example, a socratic setting of the original dataset where the step-by-step solution is paired with guided questions. However, after accounting for confidence intervals, this gain is only significant on Qwen3-4B. Comparing Baseline (t) and Adaptive (t), we make consistent gains on most models except for Qwen3-1.7B on ``hard'' (Olympiad) math benchmarks, and Qwen3-14B on ``easy'' (grade-school) benchmarks. For Qwen3-1.7B we believe the model is too small to think of any effective question, whereas on Qwen3-14B, the model is already saturated on GSM8k and more likely to get confused on easier samples.
To study $\Delta_{acc}$, the rest of our ablations only use non-thinking mode. We conduct further analysis below.

\vspace{-0.75em}
\paragraph{Question-asking vs. increased thinking.} 
Overthinking is well-studied in previous works \cite{chen2024not, sui2025stop} --- in thinking mode, Qwen3-14B uses 20x more tokens than non-thinking mode on GSM8k for only +0.6pp improvement.
However, as shown in \Cref{tab:belief-cost-per-model}, our adaptive setting achieves greater token efficiency for the gained accuracy (+4.7 points accuracy compared to the non-thinking baseline for $\approx$2x the tokens in our adaptive setting on Qwen3-1.7B, compared to thinking which is +10.7 points accuracy for $\approx$10x the tokens). 
The gain is more apparent with the Adaptive (t) setting. Our QA pipeline allows the model to complete generation faster than zero-shot prompting (10x reduction in Qwen3-4B and Qwen3-14B). While there are no dramatic performance improvements (the Confidence Interval at 30 samples is $\approx \pm$15pp), our adaptive setting observes consistent improvements.

We also observe our gating policy is weakly aligned to difficulty and supports the concept of token budgeting by naturally gating more for uncertain or weak reasoning steps. In \Cref{tab:belief-cost-per-model}, the mean number of backtracks typically increases with dataset difficulty. AIME24, our most challenging dataset, has the most mean backtracks per sample. For Math500 in our adaptive setting, the number of iterations is \emph{weakly correlated} to the level of difficulty ($\rho$ between 0.143-0.179 per model), which indicates that the samples models struggle with are only weakly correlated to human-perceived difficulty of the questions. we also find the probe score $s_t$ is even more strongly correlated to the difficulty level. This indicates that $s_t$ tracks difficulty more closely than the probe score range, even though these measures are correlated.

\vspace{-0.5em}

\begin{figure}
    \centering
    \includegraphics[width=0.7\linewidth, trim={0.6cm 0.7cm 0.6cm 1cm},clip]{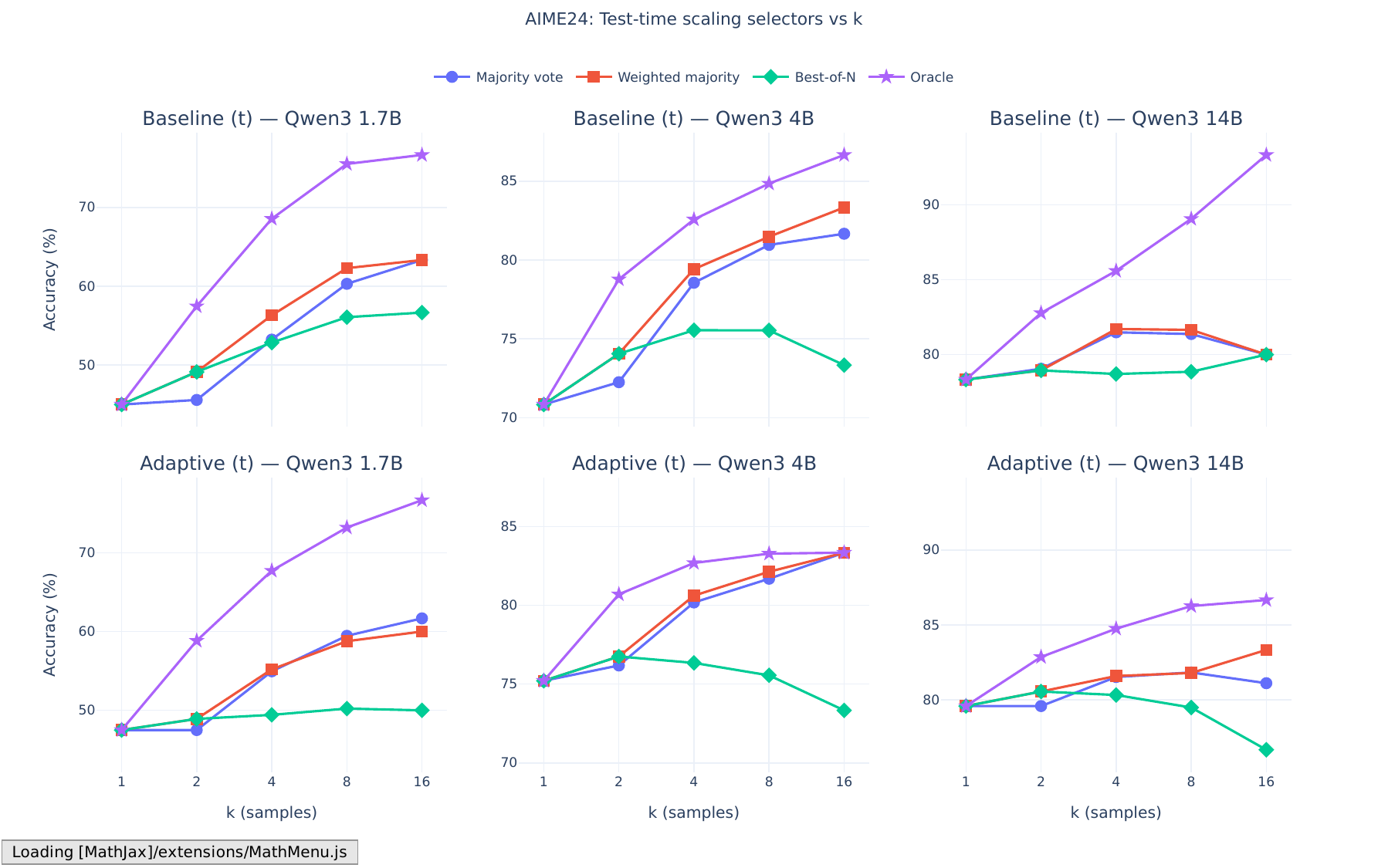}
    \caption{The effect of increasing sample counts over zero-shot and our question-asking pipeline on Qwen3 models in the AIME24 dataset.}
    \label{fig:kcurves}
\end{figure}
\paragraph{Relationship to scaling} We additionally increase the sample count $k$ in the adaptive setting to evaluate the effect of scaling, shown in \Cref{fig:kcurves}. We measured over standard metrics in test-time scaling such as weighted majority and Best-of-N (BoN), as determined by the PRM score. As shown by the Oracle setting (whether or not the correct answer was in any of the sampled trajectories), there is still a benefit to sampling more from the same model in both zero-shot and question-asking settings. It is also interesting to note that the BoN most preferred trajectory according to the PRM is often not the one preferred by majority vote, and the latter often produces a better outcome. Finally, we observe the scaling is more stable in our Question-asking setting than in the zero-shot setting. While the oracle scoring can achieve impressive scaling, especially in Qwen3-14B, the answer is not easily identified by majority vote or PRM-based scoring, indicating there is some randomness in how the trajectory is found. In contrast, the weighted majority in our question-asking setting increases monotonically. This demonstrates that there is more stability in the trajectories sampled after our question-asking pipeline, which is desirable in applications that require a consistent output.
  
\vspace{-0.5em}
\paragraph{Measuring $\hat\Delta_{acc}$ and $\Delta_{q}$ in Question-asking.} In \Cref{tab:setting-comparison}, we isolate the benefit of our question-asking intervention from pure resampling by taking the subset of all samples and models where the gating policy triggered. We define a wrong sample as one where the baseline's majority-vote accuracy was wrong, and divide the results by correct vs. incorrect at the zero-shot baseline.
The No Gating setting is where we intervene on every sample for the full $m$ iterations, where wrong samples only represent 23.3\% of the samples; belief-state gating with fixed scheduling fires on wrong samples even less frequently at 17\%. In our adaptive setting, the gating policy uses the internal hidden state probe and an external PRM; of the trajectories caught by this policy, 51.4\% were wrong in their measured baseline. This demonstrates the policy itself is efficient at catching samples that are likely wrong.

In confident baselines (all correct or all incorrect), $\delta_{acc}$ increases monotonically with selectivity ($+$3.2 to $+$8.0, then $+$19.29). 
We observe samples that were incorrect can recover to a correct answer, but our intervention can also \emph{harm} confidently correct samples; for the fixed trigger scheduling, $|\hat\Delta_{acc}|$ is \emph{greater} in magnitude on confidently correct samples than on confidently incorrect ones. While there are many samples that flip predictions in less confident bins (e.g. 136/168 had some shift in the adaptive setting, $\leq$2/5 bin), the magnitude of $\hat\Delta_{acc}$ is low because the trajectory does not shift towards the right answer. The marginal $\Delta_{acc}$ is relatively even, within 1\% pp of the baseline.

This asymmetry reflects the same mechanism identified by \citet{kumaran2025overconfidence}, where a model is more susceptible to changing its opinion when it is underconfident. Our question-asking intervention implicitly signals doubt in the current trajectory and the model treats this as necessary feedback.
Additionally, the PRM threshold filter also detects samples where, even though the trajectory is confidently correct, might contain reasoning steps where the model is less confident.

\begin{table*}[t]
\centering
\small
\caption{Comparison of belief-state gating (with PRM-adaptive vs. fixed scheduling) vs. no gating for fixed intervention intervals across all models. The No-gating setting is only reported on Math500, while other settings use all datasets. We report the number of samples (n), number of samples that flipped to a different prediction (nf), and $\hat\Delta_{acc}$. We also report the Correct (C)/ Wrong (W) samples fired, the total pool of samples, and the marginal $\Delta_{acc}$ relative to the portion fired and the entire pool.}
\label{tab:setting-comparison}
\setlength{\tabcolsep}{4pt}
\resizebox{0.7\linewidth}{!}{

\begin{tabular}{llrrrrrrrrrrrrrr} 
\toprule
& & \multicolumn{3}{c}{Adaptive, Gating} & \multicolumn{3}{c}{Fixed, Gating} & \multicolumn{3}{c}{Fixed, No gating} \\
\cmidrule(lr){3-5} \cmidrule(lr){6-8} \cmidrule(lr){9-11}
Direction & Bin & n & nf & \textbf{$\Delta$acc} & n & nf & \textbf{$\Delta$acc} & n & nf & \textbf{$\Delta$acc} \\
\midrule
\multirow{4}{*}{Correct}
 & $\le$ 2/5 &      30 &  14 & \textbf{\phantom{0}+4.7} &      60 &  29 & \textbf{\phantom{0}+1.3} &      48 &  23 & \textbf{\phantom{0}+0.0} \\
 & 3/5      &      39 &  11 & \textbf{\phantom{0}+2.6} &      88 &  30 & \textbf{\phantom{0}+2.3} &      69 &  17 & \textbf{\phantom{0}+5.8} \\
 & 4/5      &      58 &  12 & \textbf{\phantom{0}-5.5} &     137 &  22 & \textbf{\phantom{0}-7.0} &     129 &  17 & \textbf{\phantom{0}-3.2} \\
 & 5/5      &     135 &  15 & \textbf{-13.8}           &     582 &  25 & \textbf{\phantom{0}-6.4} & 1{,}084 &  13 & \textbf{\phantom{0}-2.8} \\
\midrule
\multirow{4}{*}{Wrong}
 & $\le$ 2/5 &     168 & 136 & \textbf{\phantom{0}+5.6} &     204 & 170 & \textbf{\phantom{0}+6.4} &     189 & 157 & \textbf{\phantom{0}+6.9} \\
 & 3/5      &      45 &  27 & \textbf{\phantom{0}+6.2} &      69 &  40 & \textbf{\phantom{0}+9.3} &      62 &  28 & \textbf{\phantom{0}+8.1} \\
 & 4/5      &      36 &  14 & \textbf{\phantom{0}+8.3} &      48 &  17 & \textbf{+10.0}           &      42 &  14 & \textbf{+10.5} \\
 & 5/5      &      28 &   8 & \textbf{+19.3}           &      45 &   7 & \textbf{\phantom{0}+8.0} &      63 &   8 & \textbf{\phantom{0}+3.2} \\
\midrule
\multicolumn{2}{l}{\textbf{Split (fire rate \%)}} & \multicolumn{3}{c}{\textcolor{green}{262} C / \textcolor{red}{277} W (51.4\% W)} & \multicolumn{3}{c}{\textcolor{green}{867} C / \textcolor{red}{366} W (29.7\% W)} & \multicolumn{3}{c}{\textcolor{green}{1{,}330} C / \textcolor{red}{356} W (21.1\% W)} \\
\multicolumn{2}{l}{\textbf{Pool n (fire rate \%)}}      & \multicolumn{3}{c}{3{,}839 (14.0\%)}        & \multicolumn{3}{c}{4{,}188 (29.4\%)}        & \multicolumn{3}{c}{1{,}708 (98.7\%)} \\
\multicolumn{2}{l}{\textbf{$\Delta$acc (fired)}}     & \multicolumn{3}{c}{\textbf{$+$0.22\%}}      & \multicolumn{3}{c}{$-$1.31\%}               & \multicolumn{3}{c}{$-$0.35\%} \\
\multicolumn{2}{l}{\textbf{$\Delta$acc (pool)}}      & \multicolumn{3}{c}{\textbf{$+$0.031\%}}     & \multicolumn{3}{c}{$-$0.39\%}              & \multicolumn{3}{c}{$-$0.34\%} \\
\bottomrule
\end{tabular}
}
\end{table*}

\vspace{-0.5em}

  \begin{figure}
      \centering
      \includegraphics[width=0.65\linewidth, trim={0.5cm 0.5cm 0.5cm 0cm},clip]{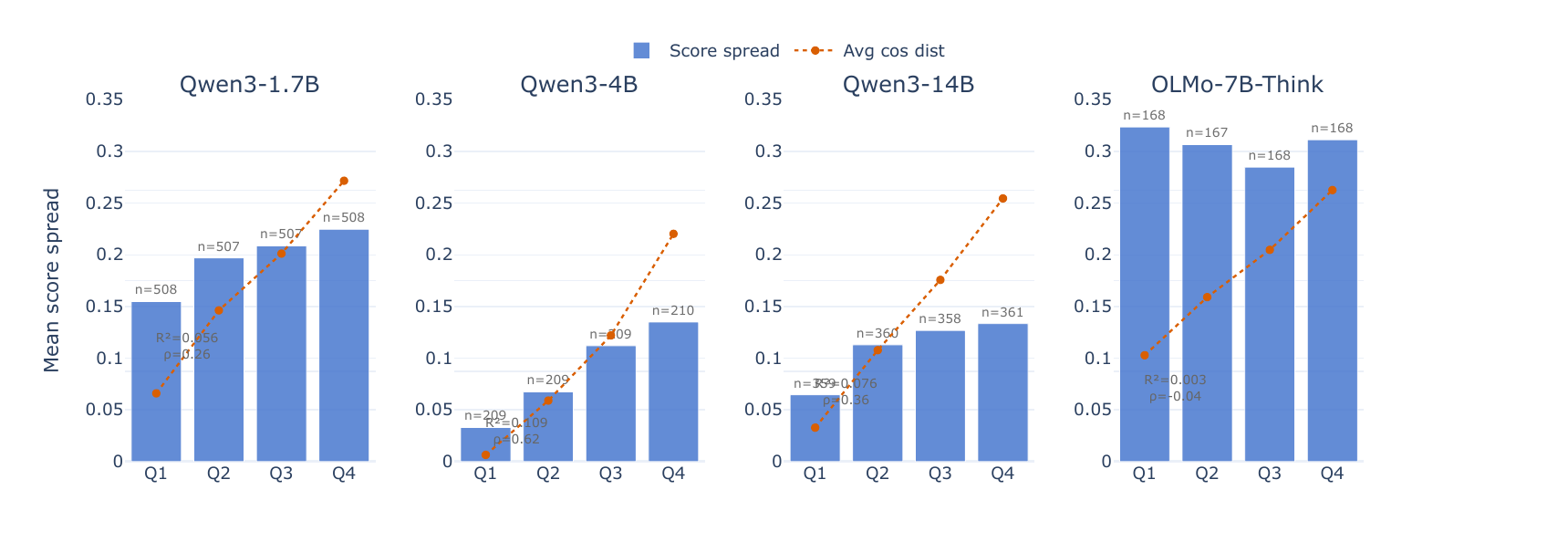}
      \caption{Correlation between semantic diversity (measured as cosine similarity over TF-IDF embeddings) versus probe score on the Adaptive setting, averaged across all datasets. The semantic diversity is divided into four quartiles and plotted against the mean range in each quartile.}
      \label{fig:diversity}
  \end{figure}
\subsection{What makes a good question?}
\paragraph{Quantifying question diversity} We measure diversity along two dimensions: sample diversity, i.e. how many of the $m=8$ questions are unique, and semantic diversity, i.e. how unique is the semantic content of the documents. We use TF-IDF embeddings for fast calculation, as shown in \Cref{fig:diversity}. There is some weak relationship between semantic diversity and score range in the Qwen3 models --- we measured $\rho$ scores between 0.258--0.619 --- but the correlation is weaker in Olmo-3-7B-Think ($\rho=-0.042$). This indicates the hidden state information learned by the probe is not strongly related to the semantics of the question.

\begin{figure}
    \centering
    \begin{subfigure}[t]{0.45\textwidth}
        
        \includegraphics[width=\linewidth , height=3.2cm, trim=1cm 0 0 3cm, clip]{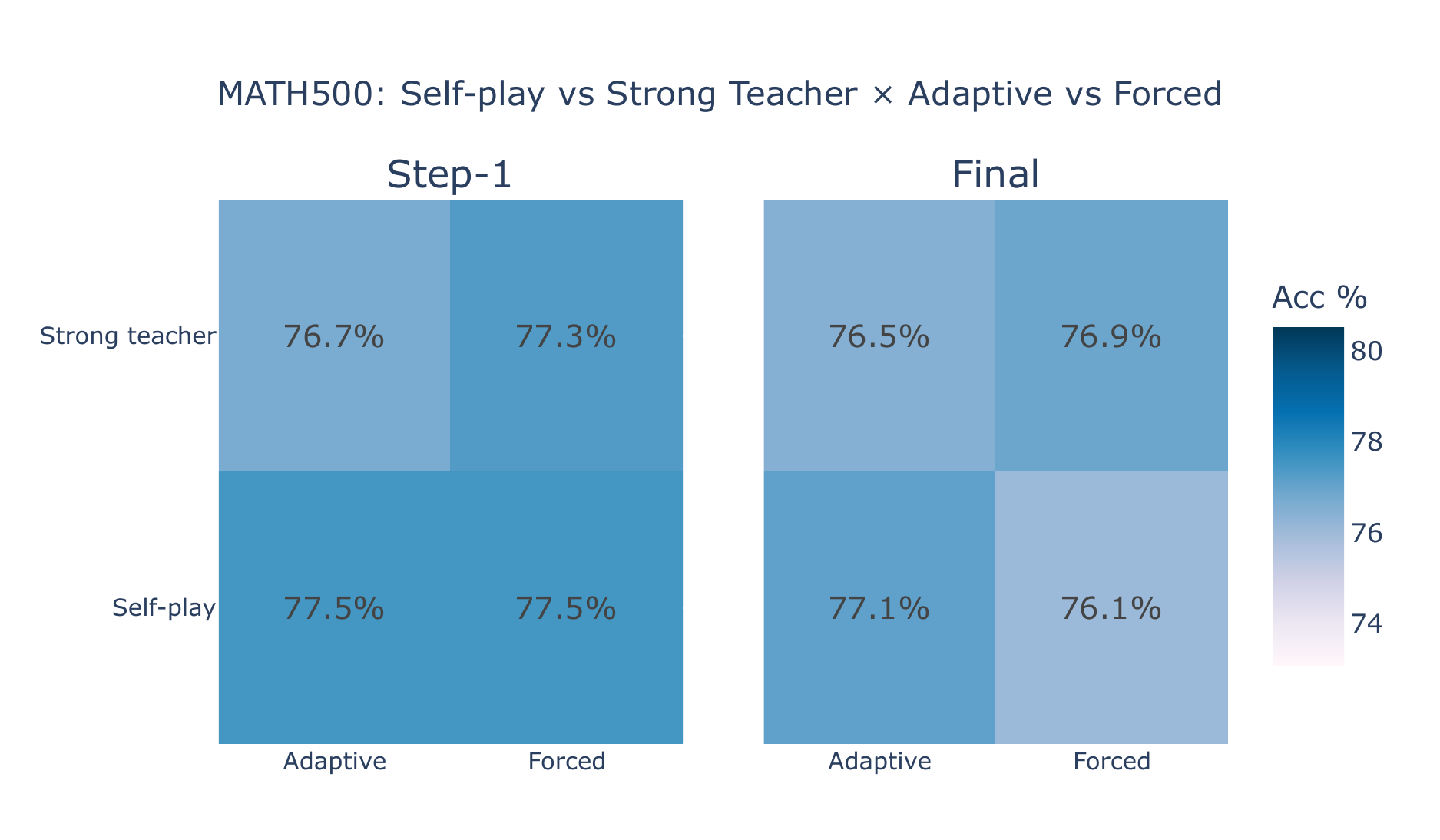}
        \caption{Math500.}
    \end{subfigure}
    \begin{subfigure}[t]{0.45\textwidth}
        
        \includegraphics[width=\linewidth , height=3.2cm, trim=1cm 0 0 3cm, clip]{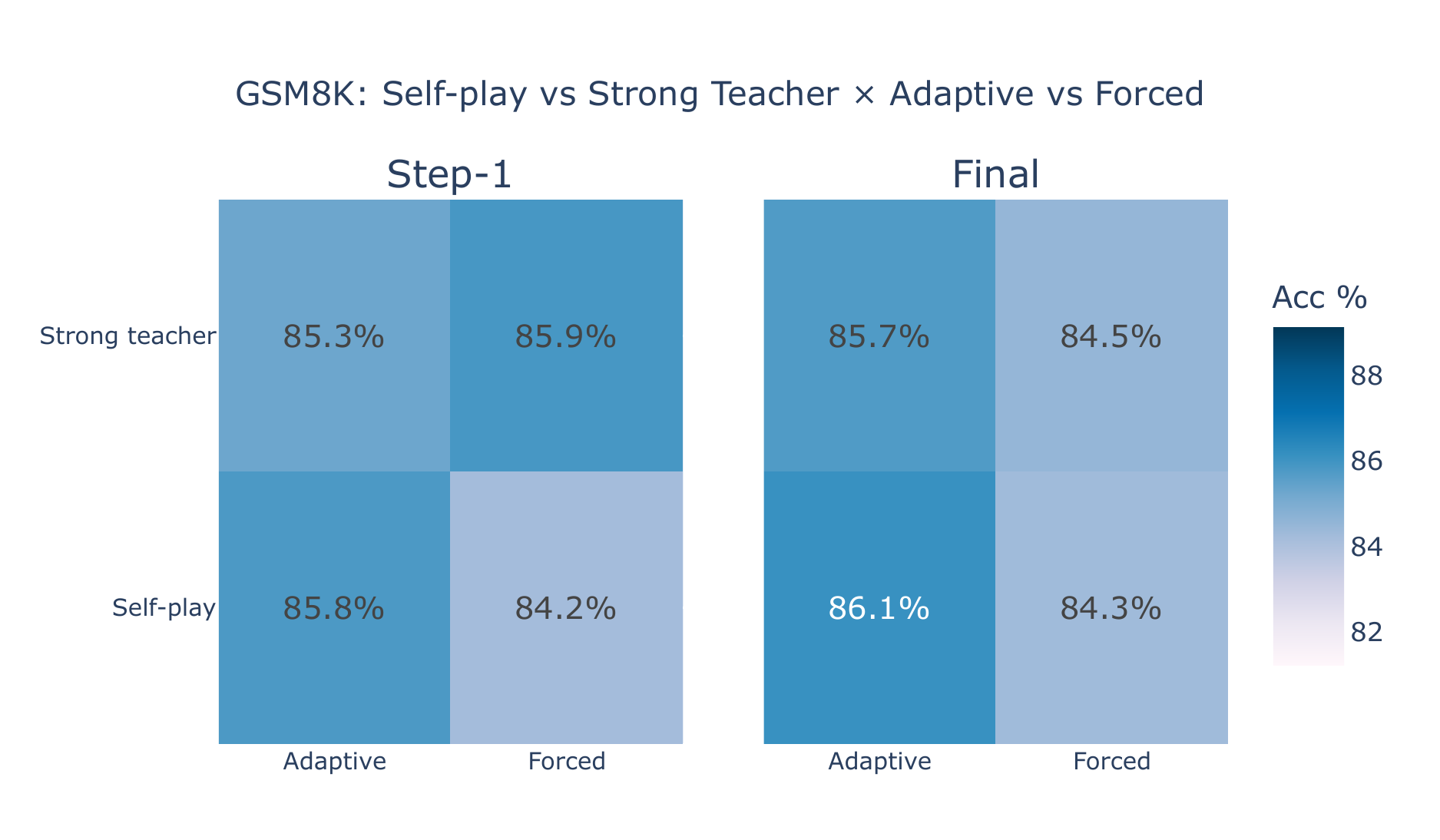}
        \caption{GSM8k.}
    \end{subfigure}
        

    \caption{Effect of teacher model and multiple question-asking iterations on Qwen3-1.7B, measured as Majority vote accuracy on Math500 and GSM8k. Self-play is where the student and teacher model share weights, and \textit{Strong teacher} is where we use Qwen3-14B as the teacher.
}
    \label{fig:teacher}
\end{figure}
\paragraph{Effect of the teacher's answer} In \Cref{fig:teacher}, we additionally analyze the effect of the teacher answer on the performance of our weakest model, Qwen3-1.7B. We compare self-play to a stronger model, Qwen3-14B, on GSM8k and Math500, and examine the dynamics of the fixed setting over early stopping at the first iteration, vs. the final output after five iterations. Without an adaptive scheduler, the fixed setting of our pipeline actively removes all benefits of the strong teacher and \textit{results in worse performance than self-play}. However, the adaptive setting slightly improves the best teacher performance. This suggests there exists an ideal timing where an intervention helps steer an otherwise incorrect trajectory towards the correct answer.
We theorize the adaptive setting has slightly more positive shifts because it is selective and allows more targeted interventions, even when we observe the PRM is poor at selecting the most correct answer in the BoN metric. 
We then use simple regex string matching to categorize different types of teacher answers. We observe four
categories from our MATH500 experiments:

\begin{table}[h]
\centering
\small
\resizebox{0.5\linewidth}{!}{
\begin{tabular}{lcc}
\toprule
Category & (t) MATH, $n=1136$ & (nt) MATH, $n=472$ \\
\midrule
Confirm & 22.0\% & 24.6\% \\
Correct & 3.3\%  & 5.7\%  \\
Hedge & 1.3\%  & 2.8\%  \\
other & 73.4\% & 66.9\% \\
\bottomrule
\end{tabular}
}
\end{table}

%% file: sections/appendix.tex
\appendix

\section*{Limitations}

\vspace{-0.5em}
\label{sec:limitations}
We would like to highlight several limitations of this study. Due to computational limitations, we only sample 5 trajectories in our core comparisons table. While we believe the principles discussed are applicable to non-verifiable domains, the contents of this study are intentionally limited to mathematical reasoning, over a model that was optimized to the test sets we report. We leave exploration of other domains, including logical reasoning, for future work.

\section{Extended Experimental Details}
\label{app:exp}
For the hidden-state probe, we use a 64-dim MLP architecture, trained with a regression objective over self-consistency for the belief state prediction. We trained each probe for 10 epochs and a learning rate of 1e-3, over offline trajectories sampled from a mix of 100 samples of GSM8k, 15 from Math500, and 15 from Big-Bench Extra Hard \cite{kazemi2025big}. We used a special data collection setting where, on the fixed intervention schedule, we sample self-play teacher answers, and five unique rollouts per question. See \Cref{tab:belief-regressor} for the validation accuracy results.

For our gating threshold, we choose $\gamma$ to be 0.6 after an initial hyperparameter sweep on the training data at the intervals [0.1, 0.3, 0.5, 0.7, 0.9]. We set $n$ to be 5 reasoning steps (split by newlines, not individual sentences), and $m$ to be 5. We experimented with $m = 15$, but find that there are minimal gains beyond 5 interventions in one reasoning trajectory. We measure the efficiency of the ``natural'' generation for a model --- in other words, we allow the model to generate until the EOS token.

All experiments are performed on a cluster of Nvidia H100 80GB GPUs. Experimental time depends on the model and experimental setting, $\approx$6-20 GPU hours per experiment run. All experiments were run on vLLM with fp16 precision, including the hidden state probe - we modified vLLM to host the hidden state and feed a dual-head classifier and inference endpoint, so the probe and the token generation shared the same weights and KV cache. For the sake of efficiency in the KV cache we reserved one GPU per model, so one experiment uses 1-3 GPUs depending on the student-teacher model configuration and the presence of a PRM. 

For Qwen3, except for the thinking baseline, we use non-thinking mode and the recommended decoding settings: temperature at 0.7, top-p at 0.9, and top-k at 20. Olmo-3-7B-Think has thinking mode on by default, and we also use the recommended sampling settings (temperature=0.6, top-p=0.95). We also include the full system prompts in \Cref{tab:system-prompts} and user message templates in \Cref{tab:user-templates}. For evaluation, we used standard math evaluation libraries available through lm-evaluation-harness\footnote{\url{https://github.com/EleutherAI/lm-evaluation-harness}}. To estimate confidence intervals on our main results we used bootstrapping --- we resample with replacement 1000 times, recalculate the majority-vote accuracy, and report the 2.5th and 97.5th percentile as the general confidence interval.

All artifacts used in this work are released in the public domain. Qwen3, Olmo-3, and AIME24\footnote{\url{https://huggingface.co/datasets/math-ai/aime24}} all use Apache 2.0 licenses, while Math500\footnote{\url{https://github.com/hendrycks/math/blob/main/LICENSE}} and GSM8k\footnote{\url{https://github.com/openai/grade-school-math/blob/master/LICENSE}} were released under MIT licenses. Coding agents were used to write code and monitor and re-submit experiments, and a human manually inspected and hand-corrected the generated code. LLMs were not used to formulate the core methodology or inform experimental decisions.

\begin{table*}[t]
\centering
\caption{System prompts used in Pipeline v5. Each prompt is assigned to a specific role in the student--teacher QA loop.}
\label{tab:system-prompts}
\small
\begin{tabular}{@{}p{3.2cm} p{9cm}@{}}
\toprule
\textbf{Prompt Name} & \textbf{Content} \\
\midrule

Default System Prompt (Qwen3 recommended setting) &
You are solving a problem step by step. Show your reasoning clearly. Separate each reasoning step with a blank line. Put your final answer within \verb|\boxed{}|. \\
\addlinespace

Continuation &
You are continuing your reasoning after receiving feedback on a previous attempt. Incorporate the feedback and continue from where you left off. Separate each reasoning step with a blank line. Put your final answer within \verb|\boxed{}|. \\
\addlinespace

Student Question &
You are working through a logical reasoning problem step by step. You have access to a knowledgeable teacher (oracle) who can answer questions about the problem's facts and rules. You are reviewing a specific step in your reasoning --- ask the teacher a question about that step that will help you answer the overall problem. DO NOT answer the question yourself. Do not ask the teacher for the final answer or whether an answer is correct --- the teacher will not respond to such questions. Ask about specific facts, rules, or intermediate steps instead. \\
\addlinespace

Teacher &
You are a knowledgeable oracle for a logical reasoning problem. Answer the student's question using ONLY the facts and rules provided. Do not speculate or add information not present in the problem. Your task is to answer any questions the student may have, but DO NOT PROVIDE THE CORRECT ANSWER DIRECTLY. If the student asks you whether an answer is correct, do not respond. Keep your responses short and concise --- address only the specific fact, rule, or intermediate step the student raised. \\
\addlinespace

Full Answer &
You have been exploring a problem step by step and received feedback from a teacher. Based on your exploration and the feedback, produce a complete solution. Show your reasoning clearly and put your final answer within \verb|\boxed{}|. \\
\addlinespace

Rewrite Step\textsuperscript{\dag} &
You are solving a problem step by step. Show your reasoning clearly. Separate each reasoning step with a blank line. Put your final answer within \verb|\boxed{}|. \newline\newline Important: \texttt{\{teacher\_response\}} \\

\bottomrule
\end{tabular}
\vspace{2pt}
\par\raggedright\footnotesize\textsuperscript{\dag}Dynamic prompt: \texttt{\{teacher\_response\}} is replaced with the teacher's answer, stated as a bare fact so the model naturally incorporates it.
\end{table*}

\begin{table*}[t]
\centering
\caption{User message templates in Pipeline v5. Placeholders in \texttt{braces} are filled at runtime. Each template is paired with the corresponding system prompt from Table~\ref{tab:system-prompts}.}
\label{tab:user-templates}
\small
\begin{tabular}{p{2.5cm} p{9.5cm}@{}}
\toprule
\textbf{System Prompt} & \textbf{User Message Content} \\
\midrule

Baseline  &
\texttt{\{task\_question\}} \\
\addlinespace

Continuation &
Problem:\newline\texttt{\{task\_question\}}\newline\newline Feedback from previous attempt:\newline\texttt{\{prior\_feedback\}}\newline\newline Continue solving the problem, incorporating the feedback above. \\
\addlinespace

Student Question &
Problem: \texttt{\{task\_question\}}\newline\newline Your reasoning so far:\newline\texttt{\{partial\_solution\}}\newline\newline The step you are reviewing:\newline\textgreater{} \texttt{\{uncertain\_step\}}\newline\newline Ask a probing question that stress-tests this step. \\
\addlinespace

Teacher &
Full problem:\newline\texttt{\{problem\_text\}}\newline\newline Student's work so far:\newline\texttt{\{partial\_solution\}}\newline\newline Student's question: \texttt{\{question\}}\newline Answer using only the facts and rules above. \\
\addlinespace

Full Answer &
Problem:\newline\texttt{\{task\_question\}}\newline\newline Your exploration so far:\newline\texttt{\{exploration\}}\newline\newline Teacher feedback:\newline Q1: \texttt{\{q\}} / A1: \texttt{\{a\}} \ldots\newline\newline Based on the exploration and feedback above, determine the answer. Show your complete reasoning and put your final answer within \verb|\boxed{}|. \\
\addlinespace

Rewrite Step &
\texttt{\{problem\_text\}}\newline\newline (Prior clean steps appended as generation prefix) \\

\bottomrule
\end{tabular}
\end{table*}







\section{Additional Experimental Results}
\label{app:extra-tabs}
\paragraph{Intervention strategies} We compare two intervention policy choices for $\pi(\cdot \mid \tau_t)$ to analyze the effect of available information on the final success:
\begin{itemize}
    \item \textbf{Backtrack:} Framing the intervening step as a ``bad'' step, we prompt the model to re-generate the step based on the teacher's feedback but \emph{without seeing the original question.}
    \item \textbf{Rewrite:} Our full QA intervention. The student has access to their original question, the teacher's answer, and they re-generate the step with full information.
\end{itemize}
\begin{figure}
    \centering
    \includegraphics[width=0.5\linewidth, trim=0 0 0 1cm, clip]{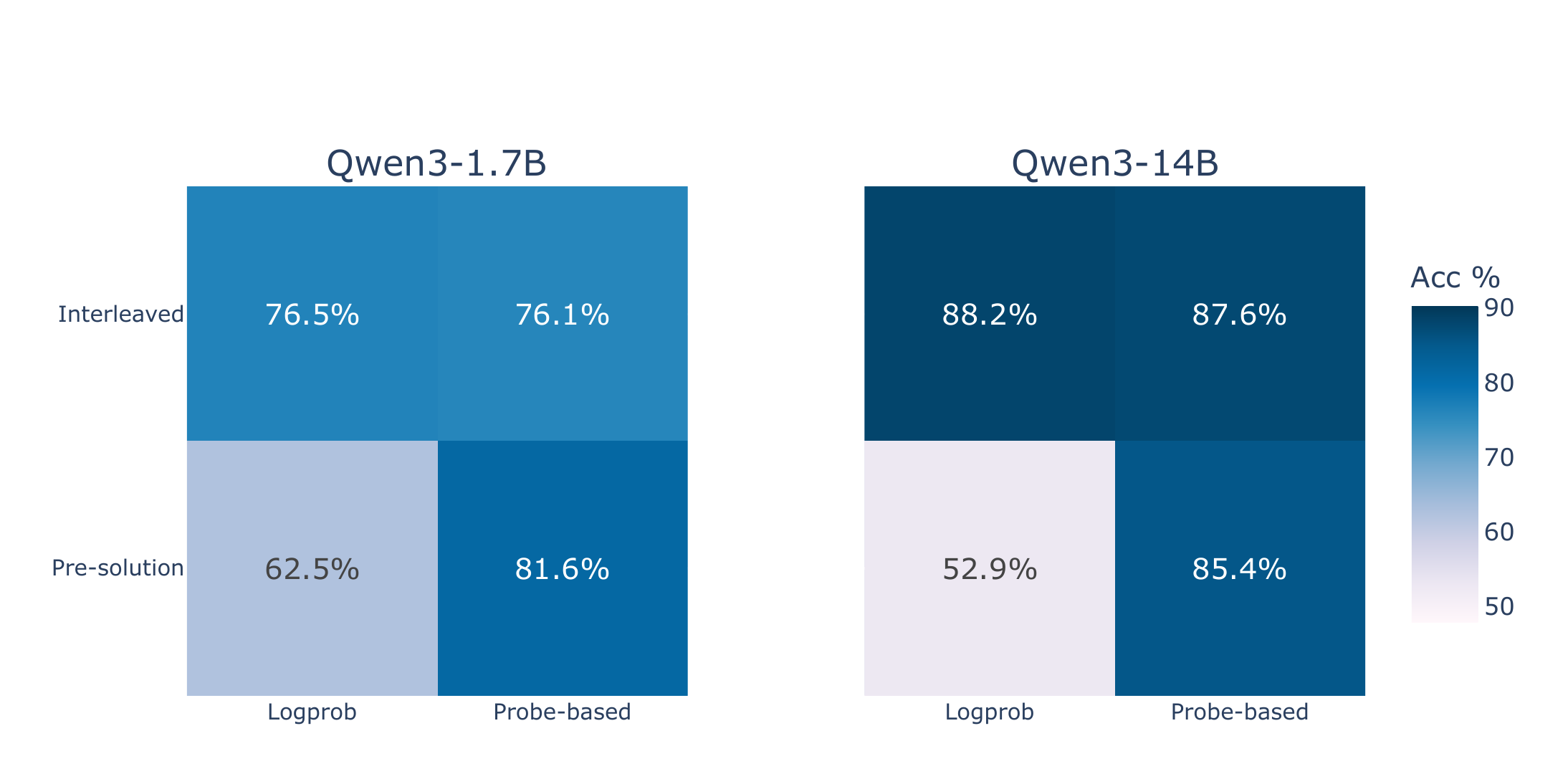}
    \caption{Comparing available model context vs. question selection on Math500 results across Qwen3-1.7B and Qwen3-14B. Pre-solution is where we ask all questions before the student attempts to solve the problem, and interleaved is where the student first generates $n$ reasoning steps before we sample a question. Finally, we compare sampling by highest logprob vs. our probe-based probable correctness scoring.}
    \label{fig:placement-selection}
\end{figure}
These results are available in \Cref{tab:rewrite-vs-backtrack}.
\begin{table}[t]                                                                 
  \centering                                                                       
  \caption{Belief-state regressor training results. The new regressor uses         
  concatenated pre-question and post-question hidden states with a cons-only       
  prediction target, compared to the old regressor which used pre-question hidden  
  states only with a dual cons + $\Delta$cons target. Variance explained is $1 -   
  \text{MSE} / \text{Var}(y)$, where $\text{Var}(y) = \bar{y}(1-\bar{y})$ is the 
  Bernoulli variance of the cons labels.}                                        
  \label{tab:belief-regressor}
  \begin{tabular}{llccccc}
  \toprule
  & Model & Pairs & Frac.\ wrong & Val MSE & Target var & Var.\ explained \\
  \midrule
  \multirow{3}{*}{\rotatebox{90}{\small Old}}                                      
  & Qwen3-1.7B & 511 & 0.881 & 0.049 & 0.122 & 60\% \\
  & Qwen3-4B   & 926 & 0.773 & 0.073 & 0.175 & 58\% \\                             
  & Qwen3-14B  &  39 & 0.949 & 0.272 & 0.049 &  0\% \\      
  \midrule                                                                         
  \multirow{3}{*}{\rotatebox{90}{\small New}}               
  & Qwen3-1.7B &  988 & 0.598 & 0.048 & 0.243 & 80\% \\                            
  & Qwen3-4B   & 1533 & 0.639 & 0.037 & 0.184 & 80\% \\                            
  & Qwen3-14B  &  556 & 0.536 & 0.113 & 0.249 & 55\% \\                            
  \bottomrule                                                                      
  \end{tabular}                                                                    
  \end{table}             

\begin{table}[t]
\centering
\small
\setlength{\tabcolsep}{5pt}
\caption{Backtrack vs.\ rewrite correction (PRM-triggered, belief-gated, adaptive). Acc and MV are accuracy of last-step answer and majority vote (\%); Bt is mean backtracks per sample; Tok is mean tokens per sample.}
\label{tab:rewrite-vs-backtrack}
\resizebox{0.7\linewidth}{!}{%
\begin{tabular}{llrrrr|rrrr}
\toprule
 & & \multicolumn{4}{c|}{Backtrack} & \multicolumn{4}{c}{Rewrite} \\
\cmidrule(lr){3-6}\cmidrule(lr){7-10}
Model & Dataset & Acc & MV & Bt & Tok & Acc & MV & Bt & Tok \\
\midrule
Qwen3-1.7B & GSM8K & 93.2 & 85.6  & 0.0 & 359 & 91.0 & 86.1  & 0.1 & 318 \\
 & MATH500 & 80.0 & 73.2  & 0.2 & 1201 & 77.5 & 72.4  & 0.5 & 1260 \\
 & AIME24 & 23.3 & 20.0  & 0.8 & 5180 & 23.3 & 13.3  & 1.2 & 4584 \\
\midrule
Qwen3-4B & GSM8K & 95.2 & 92.5  & 0.0 & 286 & 95.4 & 92.9  & 0.0 & 296 \\
 & MATH500 & 86.0 & 80.2  & 0.1 & 1152 & 90.8 & 85.6  & 0.2 & 1104 \\
 & AIME24 & 36.7 & 23.3  & 0.0 & 5439 & 33.3 & 30.0  & 1.0 & 5317 \\
\midrule
Qwen3-14B & GSM8K & 97.7 & 96.6  & 0.0 & 207 & 96.6 & 95.2  & 0.0 & 200 \\
 & MATH500 & 85.6 & 84.7  & 0.1 & 1338 & 86.0 & 83.7  & 0.2 & 979 \\
 & AIME24 & 50.0 & 30.0  & 0.0 & 4600 & 40.0 & 33.3  & 0.9 & 3915 \\
\bottomrule
\end{tabular}
}
\small

\end{table}
\subsection{Scaling on Math500}
\begin{figure}
    \centering
    \includegraphics[width=0.8\linewidth, trim={0.5cm 0.5cm 0.5cm 0cm},clip]{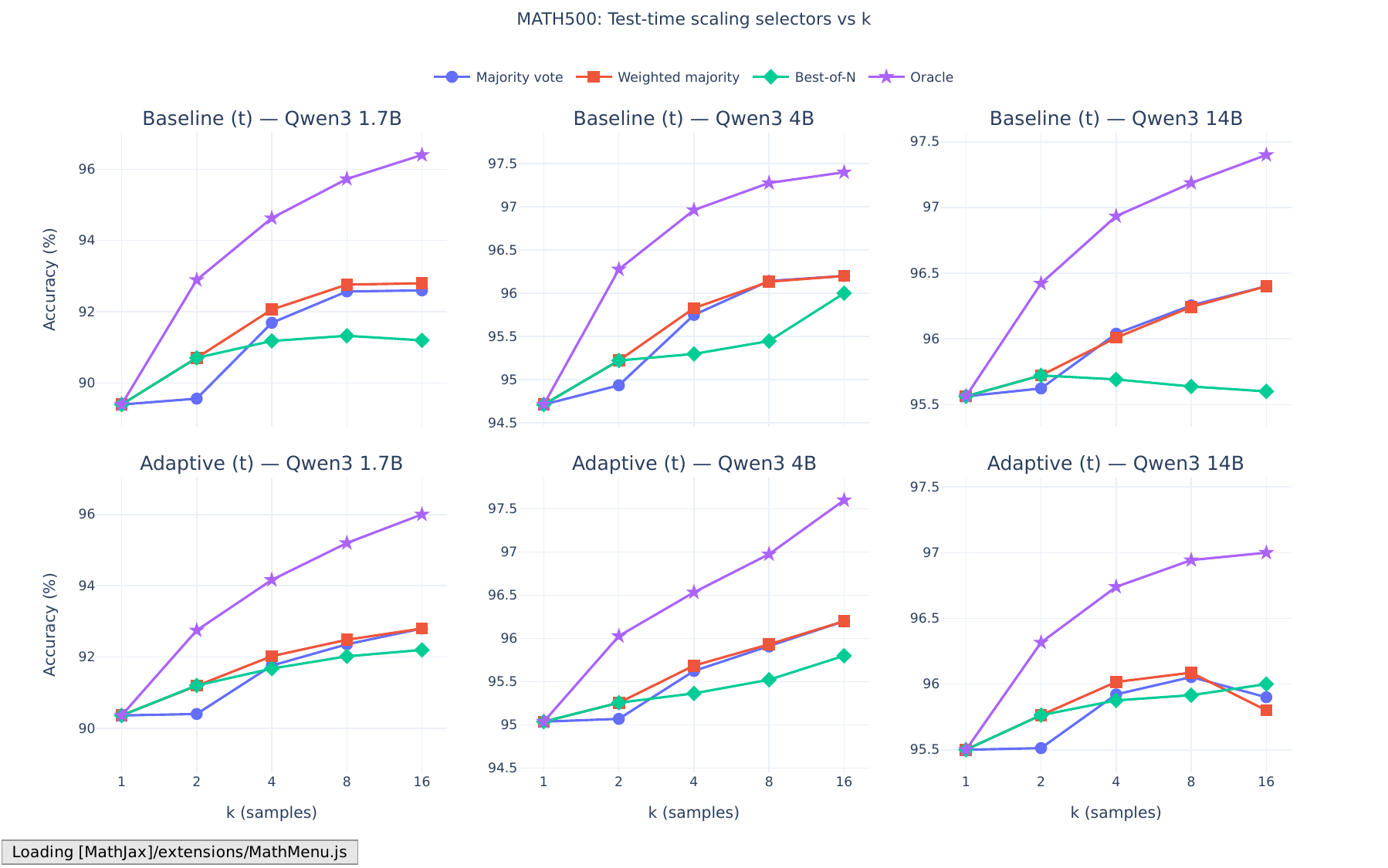}
    \caption{The effect of increasing sample counts over zero-shot and our question-asking pipeline on Qwen3 models in the AIME24 dataset.}
    \label{fig:placeholder}
\end{figure}

\subsection{Qualitative analysis of Questions}
\label{app:questions}
While we originally had some clustering analysis of the
questions, which is where the semantic diversity graphs from Figure~3 were derived from, we
omitted them from the final draft for brevity. Here are some example clusters from that
study:

\paragraph{Cluster 0 ($n=263$):}
\begin{itemize}[leftmargin=2em,itemsep=0.2em,topsep=0.2em]
  \item What if the number of junior programmers is not $2/5$ of the total number of
        employees, but instead $2/5$ of the total number of senior programmers? Would that
        change the calculation?
  \item How does the principle of inclusion-exclusion apply when we are given the total
        number of students, the number in each club, and the number in neither? What
        assumptions are being made that could potentially affect the result?
  \item What if the total number of students in the hall was not 1000, but a different
        number, say 1001? Would the reasoning still hold?
\end{itemize}

\paragraph{Cluster 1 ($n=71$):}
\begin{itemize}[leftmargin=2em,itemsep=0.2em,topsep=0.2em]
  \item What if the cost of lemons or sugar was not per gallon but per glass? Would that
        affect the calculation of total profit?
  \item \emph{Probing Question:} Assuming the ratio of sugar to water is $7{:}13$, and the
        total amount of sugar is 120 teaspoons, does the calculation
        correctly account for the fact that the amount of water is in cups, not teaspoons?
        What if the conversion from cups to teaspoons is not fixed or known?
\end{itemize}

\paragraph{Cluster 2 ($n=641$):}
\begin{itemize}[leftmargin=2em,itemsep=0.2em,topsep=0.2em]
  \item Is there any possibility that the fraction $2/5$ refers to something other than the
        proportion of boys among the total Grade 5 students, and could this affect the
        calculation?
  \item Is there any possibility that the first 3 customers did not each buy exactly 1 DVD,
        and if so, how would that affect the calculation?
\end{itemize}

\paragraph{Cluster 3 ($n=62$):}
\begin{itemize}[leftmargin=2em,itemsep=0.2em,topsep=0.2em]
  \item \emph{Question:} Is there a possibility that the assumption that
        is isosceles with
        leads to an incorrect conclusion about the angle
        , and does this affect the validity of the final answer?
  \item Is there a guarantee that the given side lengths
        satisfy the triangle inequality, ensuring that the triangle (and hence the
        tetrahedron) can exist?
\end{itemize}

\paragraph{Cluster 4 ($n=166$):}
\begin{itemize}[leftmargin=2em,itemsep=0.2em,topsep=0.2em]
  \item What if there were more than one type of mutant invertebrate, each with a different
        number of legs? Would the calculation still hold?
  \item What if there were more than one type of mutant invertebrate, and some had 8 legs
        while others had 6? Would the calculation still hold?
\end{itemize}

\paragraph{Cluster 5 ($n=139$):}
\begin{itemize}[leftmargin=2em,itemsep=0.2em,topsep=0.2em]
  \item Is the ``half as long'' referring to half the time he spends watching TV, or half the
        total time spent on both TV and reading in each session?
  \item Is there a possibility that ``half as long'' could refer to half the total time spent
        on both TV and reading in a single session, rather than half the time spent watching
        TV?
\end{itemize}

\paragraph{Cluster 6 ($n=166$):}
\begin{itemize}[leftmargin=2em,itemsep=0.2em,topsep=0.2em]
  \item \emph{Question:} Given the expression
        , is it possible that the minimum value of this expression is 100? Why or why not?
\end{itemize}

\begin{table}[t]
\renewcommand\arraystretch{0.95}


    \centering
    \setlength{\tabcolsep}{8pt}
  \caption{Accuracy (\%) across models, datasets, and intervention strategies compared against a zero-shot baseline. The best results are in \textbf{bold}, and \textit{strong} results are within 1 point of accuracy of the best performance. --- indicates results are not available as there is no equivalent non-thinking setting for Olmo-3-Think. $^\star$ indicates we obtained the value from \cite{yang2025qwen3}}
  \label{tab:accuracy}

\resizebox{0.9\linewidth}{!}{%
\begin{tabular}{cllcc|ccc|ccc}
  \toprule
  & & &\multicolumn{2}{c|}{Baseline}   & \multicolumn{3}{c|}{Front-loaded}  &  \multicolumn{3}{c}{Interleaved}\\
  & & & non-thinking & thinking & logprob & probed     & random
   & Fixed & Adaptive & Adaptive (t) \\
  \midrule
  \multirow{6}{*}{\rotatebox{90}{Qwen 3-1.7B}}
    & \multirow{2}{*}{GSM8K}
     & P@5  & 92.6 & \textbf{95.2} & 25.2 & 88.0 & 78.8 & 87.9 & 91.0 & \textit{94.2} \\
    &  & Maj  & 84.8 & \textit{91.5} & 24.6 & 85.5 & 76.3 & 84.2 & 86.1 & \textbf{92.2} \\
    \cmidrule(l){2-11}
    & \multirow{2}{*}{MATH500}
     & P@5  & 81.6 & \textit{95.2} & 67.6 & 77.9 & 70.9 & 76.5 & 75.9 & \textbf{96.0} \\
    &  & Maj  & 74.8 & \textbf{92.8} & 59.6 & 76.3 & 61.9 & 70.1 & 70.9 & \textit{92.6} \\
    \cmidrule(l){2-11}
    & \multirow{2}{*}{AIME24}
     & P@5  & 30.0 & 48.3$^\star$ & 10.0 & -- & 0.0 & 20.0 & 23.3 & \textbf{66.7 }\\
    &  & Maj  & 20.0 & 48.3$^\star$ & 10.0 & -- & 0.0 & 13.3 & 13.3 & \textbf{53.3} \\
  \midrule
  \multirow{6}{*}{\rotatebox{90}{Qwen 3-4B}}
    & \multirow{2}{*}{GSM8K}
     & P@5  & \textit{96.7} & \textit{96.7} & 69.9 & 93.7 & 46.1 & \textbf{97.4} & 95.4 & \textit{96.9} \\
    &  & Maj  & 92.7 & \textit{95.2} & 61.3 & 90.3 & 42.2 & \textit{95.3} & 93.0 & \textbf{96.1} \\
    \cmidrule(l){2-11}
    & \multirow{2}{*}{MATH500}
     & P@5  & 87.6 & \textit{98.0} & 43.1 & 80.8 & 12.6 & 83.1 & 82.6 & \textbf{98.5} \\
    &  & Maj  & 82.1 & 96.7 & 34.6 & 73.6 & 10.9 & 77.8 & 77.6 & \textbf{98.5} \\
    \cmidrule(l){2-11}
    & \multirow{2}{*}{AIME24}
     & P@5  & 33.3 & \textbf{80.0} & 66.7 & -- & 64.3 & 39.3 & 33.3 & \textbf{80.0} \\
    &  & Maj  & 33.3 & 76.7 & 60.0 & -- & 57.1 & 28.6 & 30.0 & \textbf{80.0} \\
  \midrule
  \multirow{6}{*}{\rotatebox{90}{Qwen 3-14B}}
    & \multirow{2}{*}{GSM8K}
     & P@5  & 95.0 & \textit{97.6} & 51.0 & 95.0 & -- & 95.8 & 96.6 & \textbf{97.9} \\
    &  & Maj  & 91.6 & \textbf{96.6} & 47.2 & 91.6 & -- & 95.4 & 95.2 & \textbf{96.6} \\
    \cmidrule(l){2-11}
    & \multirow{2}{*}{MATH500}
     & P@5  & 88.8 & \textbf{98.8} & 59.9 & 84.3 & 12.8 & 86.9 & 83.9 & \textit{98.5} \\
    &  & Maj  & 87.0 & \textit{97.1} & 50.6 & 79.8 & 12.0 & 81.9 & 81.4 & \textbf{97.3} \\
    \cmidrule(l){2-11}
    & \multirow{2}{*}{AIME24}
     & P@5  & 46.4 & 80.0 & 66.7 & -- & -- & 50.0 & 40.0 & \textbf{90.0} \\
    &  & Maj  & 32.1 & 80.0 & 61.9 & -- & -- & 40.0 & 33.3 & \textbf{86.7} \\

  \midrule

  \multirow{6}{*}{\rotatebox{90}{Olmo3-Think}}
    & \multirow{2}{*}{GSM8K}
     & P@5  & -- & \textbf{98.0} & -- & --& -- & 96.9  & --& \textit{97.7} \\
    &  & Maj  & -- & \textit{94.9} & -- & -- & -- & \textbf{95.7}& -- & \textit{95.5}  \\
    \cmidrule(l){2-11}
    & \multirow{2}{*}{MATH500}
     & P@5  & -- & 94.3 & -- & -- & -- & \textbf{98.3}& -- & \textit{98.0}  \\
    &  & Maj  & -- & 91.2 & -- & -- & -- & \textit{97.2} & -- & \textbf{97.6} \\
    \cmidrule(l){2-11}
    & \multirow{2}{*}{AIME24}
     & P@5  & -- & 76.7 & -- & -- & -- & \textbf{86.7}& -- & 83.3  \\
    &  & Maj  & -- & \textbf{76.7} & -- & -- & -- & 73.3  & --& \textbf{76.7} \\
  \bottomrule
  \end{tabular}%
  }
  \end{table}

\begin{table}[t]                                                                                       
  \centering                                                
  \caption{Spearman and Pearson correlation between MATH500 difficulty (level 1--5) and number of        
  student--teacher iterations under the adaptive rewrite setting (\texttt{v4\_prm\_belief\_rewrite}).}     
  \label{tab:difficulty-iterations-corr}                                                                 
  \small                                                                                                   \begin{tabular}{@{}l rr rr r@{}}
  \toprule
  \textbf{Model} & \textbf{Spearman $\rho$} & \textbf{$p$} & \textbf{Pearson $r$} &
  \textbf{$p$} & \textbf{Mean iter.} \\
  \midrule
  Qwen3-1.7B       & 0.143 & $1.3\!\times\!10^{-3}$ & 0.132 & $3.1\!\times\!10^{-3}$ & 1.23 \\
  Qwen3-4B        & 0.165 & $2.2\!\times\!10^{-4}$ & 0.135 & $2.5\!\times\!10^{-3}$ & 1.10 \\
  Qwen3-14B        & 0.137 & $2.2\!\times\!10^{-3}$ & 0.084 & $6.0\!\times\!10^{-2}$ & 1.15 \\
  Olmo-3-7B-Think  & 0.179 & $5.8\!\times\!10^{-3}$ & 0.149 & $2.2\!\times\!10^{-2}$ & 1.07 \\
  \bottomrule
  \end{tabular}
  \vspace{2pt}
  {\footnotesize Difficulty parsed from the MATH500 \texttt{level} field. All four models show a small but consistent positive correlation: harder
  problems trigger slightly more iterations.}
  \end{table}


\begin{table}[]
    \centering
    \caption{Calibration scores for the hidden state belief probe against correctness rate across three Qwen3 model sizes. *** indicates p<0.001.}
    \label{tab:calib}
    \setlength{\tabcolsep}{5pt}
    \resizebox{0.6\linewidth}{!}{
\begin{tabular}{lrcccc}
\toprule
Model & $n$ & $\rho$ (cons.) & $\rho$ (correct) & Brier & ECE \\
\midrule
Olmo-3-7B-Think & 532 & +0.203$^{***}$ & +0.205$^{***}$ & 0.177 & 0.109 \\
Qwen3-1.7B & 1562 & +0.213$^{***}$ & +0.247$^{***}$ & 0.197 & 0.103 \\
Qwen3-4B & 1767 & +0.269$^{***}$ & +0.237$^{***}$ & 0.209 & 0.157 \\
Qwen3-14B & 857 & +0.362$^{***}$ & +0.331$^{***}$ & 0.195 & 0.154 \\
\bottomrule
\end{tabular}
}
\end{table}

\begin{table}[t]
\centering
\caption{Spearman correlation between MATH500 difficulty (level 1--5) and the belief-state probe ($\mathrm{cons}$) score range across candidate questions, comparing adaptive vs. forced budgets.
\textit{Mean step range}: per-sample mean of $(\max\mathrm{cons} - \min\mathrm{cons})$ across gated steps.
\textit{Trajectory range}: per-sample $(\max_t \max\mathrm{cons}_t) - (\min_t \min\mathrm{cons}_t)$ over all gated steps.}
\label{tab:difficulty-probe-range-corr}
\small
\begin{tabular}{@{}l rr rr rr rr@{}}
\toprule
& \multicolumn{4}{c}{\textbf{Adaptive}} & \multicolumn{4}{c}{\textbf{Forced}} \\
\cmidrule(lr){2-5} \cmidrule(lr){6-9}
& \multicolumn{2}{c}{Mean step} & \multicolumn{2}{c}{Trajectory} & \multicolumn{2}{c}{Mean step} & \multicolumn{2}{c}{Trajectory} \\
\cmidrule(lr){2-3} \cmidrule(lr){4-5} \cmidrule(lr){6-7} \cmidrule(lr){8-9}
\textbf{Model} & $\rho$ & $p$ & $\rho$ & $p$ & $\rho$ & $p$ & $\rho$ & $p$ \\
\midrule
Qwen3-1.7B       & 0.178 & $1.4\!\times\!10^{-3}$ & 0.171 & $2.2\!\times\!10^{-3}$ & 0.359 & $8.9\!\times\!10^{-12}$ & 0.377 & $6.7\!\times\!10^{-13}$ \\
Qwen3-4B         & 0.182 & $1.6\!\times\!10^{-3}$ & 0.232 & $5.5\!\times\!10^{-5}$ & 0.347 & $1.6\!\times\!10^{-12}$ & 0.366 & $8.0\!\times\!10^{-14}$ \\
Qwen3-14B        & 0.209 & $6.2\!\times\!10^{-4}$ & 0.201 & $1.0\!\times\!10^{-3}$ & 0.463 & $1.2\!\times\!10^{-6}$  & 0.477 & $5.1\!\times\!10^{-7}$  \\
Olmo-3-7B-Think  & 0.193 & $5.3\!\times\!10^{-3}$ & 0.225 & $1.1\!\times\!10^{-3}$ & 0.232 & $1.6\!\times\!10^{-2}$  & 0.221 & $2.2\!\times\!10^{-2}$  \\
\bottomrule
\end{tabular}
\end{table}